%% file: root.tex
\documentclass[letterpaper, 10 pt, conference]{ieeeconf}  

\IEEEoverridecommandlockouts                              

\usepackage{times} 
\usepackage{amsmath} 
\usepackage{amssymb}  
\usepackage{graphicx}
\usepackage{adjustbox}
\usepackage{booktabs}
\usepackage[margin=1in]{geometry}
\usepackage{xcolor}
\usepackage[hidelinks]{hyperref} 
\definecolor{linkblue}{RGB}{0,90,180} 
\usepackage[font=footnotesize]{caption}   %
\usepackage[font=footnotesize]{subcaption} 
\usepackage{cite} 

\title{\LARGE \bf
SCREP: Scene Coordinate Regression and Evidential Learning-based Perception-Aware Trajectory Generation
}

\author{Juyeop Han$^{1}$, Lukas Lao Beyer$^{1}$, Guilherme V. Cavalheiro$^{1}$, and Sertac Karaman$^{1}$
\thanks{$^{1}$The authors are with Laboratory for Information and Decision Systems (LIDS), Massachusetts Institute of Technology, Cambridge, MA 02139, USA
        {\tt\small \{juyeop, llb, guivenca, sertac\}@mit.edu}}%
}

\newcommand{\rev}{\textcolor{black}}
\newcommand{\best}[1]{\textcolor{red}{{\bfseries #1}}}
\newcommand{\second}[1]{\textcolor{blue}{\underline{#1}}}

\begin{document}

\maketitle
\thispagestyle{empty}
\pagestyle{empty}

\begin{abstract}
\label{sec:abstract}
\input{section/abstract.tex}

\end{abstract}


\section{Introduction}
\label{sec:introduction}

\input{section/introduction.tex}

\section{Related Work}
\label{sec:related_work}
\input{section/related_work.tex}

\section{System Overview}
\label{sec:overview}
\input{section/prob_and_overview.tex}

\section{Methodology}
\label{sec:methodology}

\input{section/method.tex}

\begin{table*}[t]
	\centering
	\caption{Localization accuracy under the four yaw strategies, \emph{Forward}, \emph{Vanilla}, \emph{FIM}, and our proposed method (\emph{Ours}), reported as translation\,(cm)\,/\,rotation\,($^{\circ}$). Trajectory names are abbreviated as clov\,(clover), patr\,(patrick), thri\,(thrice), wint\,(winter), amp\,(ampersand), and half\,(halfmoon), while dice and sid are unchanged; Avg denotes the average. Best results are highlighted in \best{bold}, and second-best results are highlighted in \second{underline}.}
	\label{tab:comparison_loc}
	
	\begin{subtable}[t]{0.49\textwidth}
		\centering
		\footnotesize
		\setlength{\tabcolsep}{2pt}
		\begin{adjustbox}{max width=\linewidth}
			\begin{tabular}{c||cc|cc|cc|cc}
				\toprule
				 & \multicolumn{2}{c|}{Forward} & \multicolumn{2}{c|}{Vanilla} & \multicolumn{2}{c|}{FIM} & \multicolumn{2}{c}{Ours} \\
				& Median$\downarrow$ & Mean$\downarrow$ & Median$\downarrow$ & Mean$\downarrow$ & Median$\downarrow$ & Mean$\downarrow$ & Median$\downarrow$ & Mean$\downarrow$ \\ \midrule
				clov & \second{8.0}/0.6 & 9.7/0.9 & 8.1/\best{0.4} & 8.9/\second{0.5} & 8.2/\best{0.4} & \second{8.6}/\best{0.4} & \best{7.2}/0.5 & \best{7.5}/\second{0.5} \\
				patr & 8.8/0.7 & 10.6/0.8 & 7.9/\best{0.5} & 9.3/\best{0.5} & \best{5.8}/0.7 & \best{7.9}/0.7 & \second{7.7}/\best{0.5} & \second{8.9}/\second{0.6} \\
				thri & 8.2/0.6 & \second{8.8}/0.7 & \best{8.0}/\best{0.5} & \second{8.8}/\best{0.5} & 8.2/\best{0.5} & 9.1/\best{0.5} & \best{8.0}/\best{0.5} & \best{8.6}/\best{0.5} \\
				wint & 8.4/0.7 & 9.7/0.9 & 8.6/\best{0.5} & 10.1/\best{0.5} & \best{8.3}/\best{0.5} & \second{9.5}/\best{0.5} & \best{8.3}/\best{0.5} & \best{9.1}/0.6 \\ \midrule
				amp & \second{12.8}/1.7 & \second{13.8}/1.8 & 14.3/\best{1.3} & 16.3/\best{1.3} & 14.1/\best{1.3} & 17.5/\second{1.5} & \best{11.3}/1.6 & \best{11.7}/1.8 \\
				dice & \best{13.0}/2.2 & 22.3/2.9 & 16.9/\best{1.7} & \second{20.4}/\best{1.8} & 19.1/1.9 & 26.0/\second{1.9} & \second{13.9}/\best{1.7} & \best{17.8}/\second{1.9} \\
				half & \second{11.4}/\second{2.1} & 15.9/\second{2.2} & 11.5/2.2 & \best{12.5}/2.3 & \best{11.2}/2.6 & \second{12.6}/2.5 & 12.4/\best{1.9} & 13.6/\best{1.9} \\
				sid & \second{21.9}/2.8 & 30.2/3.6 & \best{18.7}/2.1 & \best{22.9}/\second{2.0} & 32.1/\second{2.0} & 44.4/2.1 & 26.3/\best{1.7} & \second{30.1}/\best{1.8} \\ \midrule
				Avg & 10.1/1.2 & 15.1/1.7 & 10.1/\best{0.9} & \second{13.7}/\best{1.2} & \best{9.7}/\second{1.0} & 16.7/\best{1.2} & \second{9.8}/1.1 & \best{13.4}/\best{1.2} \\ \bottomrule
			\end{tabular}
		\end{adjustbox}
		\subcaption{Raw SCR output through PnP-RANSAC (\emph{SCR}).}
		\label{tab:comparison_scr}
	\end{subtable}
	\hfill
	\begin{subtable}[t]{0.49\textwidth}
		\centering
		\footnotesize
		\setlength{\tabcolsep}{2pt}
		\begin{adjustbox}{max width=\linewidth}
			\begin{tabular}{c||cc|cc|cc|cc}
				\toprule
				& \multicolumn{2}{c|}{Forward} & \multicolumn{2}{c|}{Vanilla} & \multicolumn{2}{c|}{FIM} & \multicolumn{2}{c}{Ours} \\
				& RMSE$\downarrow$ & Mean$\downarrow$ & RMSE$\downarrow$ & Mean$\downarrow$ & RMSE$\downarrow$ & Mean$\downarrow$ & RMSE$\downarrow$ & Mean$\downarrow$ \\ \midrule
				clov & 18.7/4.0 & 16.5/3.6 & 17.7/\best{2.3} & 16.3/\best{2.0} & \second{17.0}/\second{3.1} & \second{15.3}/\second{2.7} & \best{16.1}/4.3 & \best{14.8}/3.7 \\
				patr & 18.0/10.5 & \second{15.9}/8.7 & 18.5/6.9 & 16.6/6.2 & \second{17.9}/\second{5.7} & \second{15.9}/\second{4.6} & \best{16.9}/\best{1.2} & \best{15.5}/\best{1.1} \\
				thri & 18.7/\second{4.7} & 17.4/\second{3.8} & \best{15.8}/5.1 & \best{14.5}/4.6 & 18.4/7.3 & 16.7/5.8 & \second{18.2}/\best{3.3} & \second{15.9}/\best{2.9} \\
				wint & 21.4/3.0 & 19.6/2.7 & \best{17.2}/\second{2.0} & \best{15.5}/\second{1.8} & 23.1/3.4 & 21.1/2.7 & \second{18.8}/\best{1.4} & \second{16.6}/\best{1.3} \\ \midrule
				amp & \second{20.6}/\second{1.7} & \second{19.1}/\second{1.6} & 25.5/8.6 & 23.2/7.8 & 26.1/5.1 & 23.3/3.9 & \best{18.0}/\best{1.0} & \best{16.6}/\best{0.9} \\
				dice & 39.7/\second{2.3} & 26.9/\second{2.0} & \second{29.5}/2.4 & \second{25.8}/2.1 & 37.8/\best{1.7} & 31.6/\best{1.5} & \best{24.3}/3.8 & \best{21.7}/3.5 \\
				half & 23.7/2.8 & 20.4/2.3 & \second{21.8}/\best{1.6} & \second{19.5}/\best{1.4} & 23.1/\second{2.6} & 21.3/\second{2.2} & \best{18.0}/4.8 & \best{17.1}/4.4 \\
				sid & 41.8/\second{1.8} & \second{33.9}/\second{1.6} & \best{34.8}/6.6 & \best{30.9}/5.0 & 63.2/6.5 & 47.9/5.6 & \second{40.5}/\best{1.5} & 35.9/\best{1.4} \\ \midrule
				Avg & 25.3/\second{3.9} & 21.2/\second{3.3} & \second{22.6}/4.4 & \second{20.3}/3.9 & 28.1/4.6 & 23.9/3.7 & \best{21.5}/\best{2.7} & \best{19.2}/\best{2.4} \\ \bottomrule
			\end{tabular}
		\end{adjustbox}
		\subcaption{Fixed-lag smoother output (\emph{IMU+SCR}).}
		\label{tab:comparison_imu}
	\end{subtable}
\end{table*}

\section{Experiments}
\label{sec:experiments}
\input{section/experiments.tex}

\section{Conclusion}
\label{sec:conclusion}
\input{section/conclusion.tex}





\section*{ACKNOWLEDGMENTS}
We used OpenAI ChatGPT and Anthropic Claude for code design and debugging related to the experiments section, and for polishing and improving the clarity of writing across the entire paper. The authors reviewed, verified, and take full responsibility for all content.

\bibliographystyle{unsrt}
\bibliography{reference}

\end{document}

%% file: section/abstract.tex
Autonomous flight in GPS-denied indoor spaces requires trajectories that keep visual-localization error tightly bounded across varied missions.
Map-based visual localization methods such as feature matching require computationally intensive map reconstruction and have feature-storage scalability issues, especially for large environments. Scene coordinate regression~(SCR) provides an efficient learning-based alternative that directly predicts 3D coordinates for every pixel, enabling absolute pose estimation suited to onboard robotics.
We present a perception-aware trajectory planner that couples an evidential learning-based SCR pose estimator with a receding-horizon trajectory optimizer.
The optimizer steers the onboard camera toward reliable scene coordinates with low uncertainty, while a fixed-lag smoother fuses the low-rate SCR pose estimates with high-rate IMU data to provide a high-quality, high-rate pose estimate.
In simulation, our planner reduces translation and rotation RMSE by at least 4.9\% and 30.8\% relative to baselines, respectively.
Hardware-in-the-loop experiments validate the feasibility of our proposed trajectory planner under close-to-real deployment conditions.

%% file: section/introduction.tex
Vision-based localization has advanced rapidly, enabling unmanned aerial vehicle~(UAV) navigation in GPS-denied indoor environments.
Map-based visual localization leverages prior environmental knowledge for drift-free pose estimates, making localization-aware trajectory planning crucial for reliable navigation in such environments.
Most perception-aware trajectory planning approaches~\cite{PAMPC18IROS,Zhang18ICRA,Murali19ACC, Bartolomei20IROS, Spasojevic20ICRA, Zhang2020Arxiv, Kim21IROS, DiGiammarino24ECCV,APACE2024ICRA,Yu2025ArXiv, Kim25ICRA} steer the camera toward feature-rich regions to improve the pose estimate obtained by structure from motion (SfM) or visual-inertial odometry (VIO).
However, SfM-based feature maps are costly to reconstruct and scale poorly to large environments, and VIO-based methods remain limited by cumulative error even with perception-aware guidance.

\begin{figure}[tb!]
	\centering
	\begin{subfigure}[t]{\columnwidth}
		\centering
		\includegraphics[width=0.75\columnwidth]{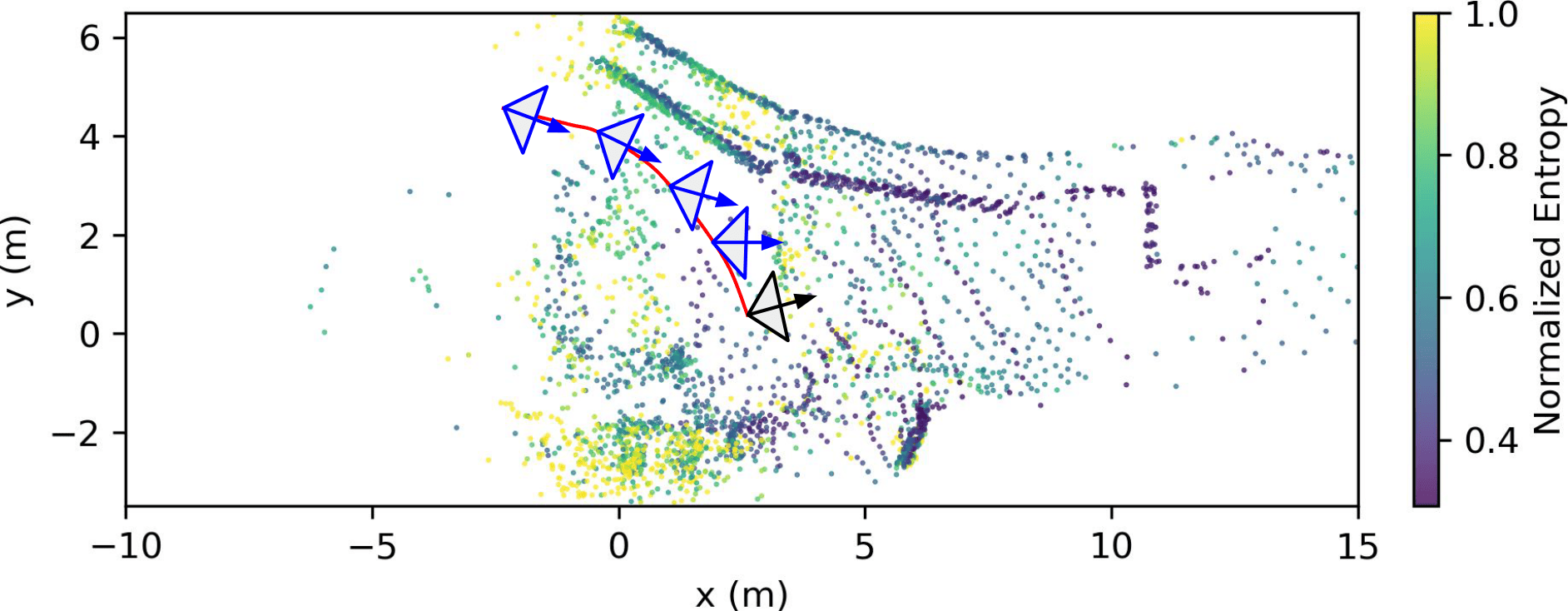}
		\caption{Illustration of the receding-horizon perception-aware trajectory generation.}
		\label{fig:hook}
	\end{subfigure}

	\vspace{1mm}
	\begin{subfigure}[t]{\columnwidth}
		\centering
		\includegraphics[width=0.9\columnwidth]{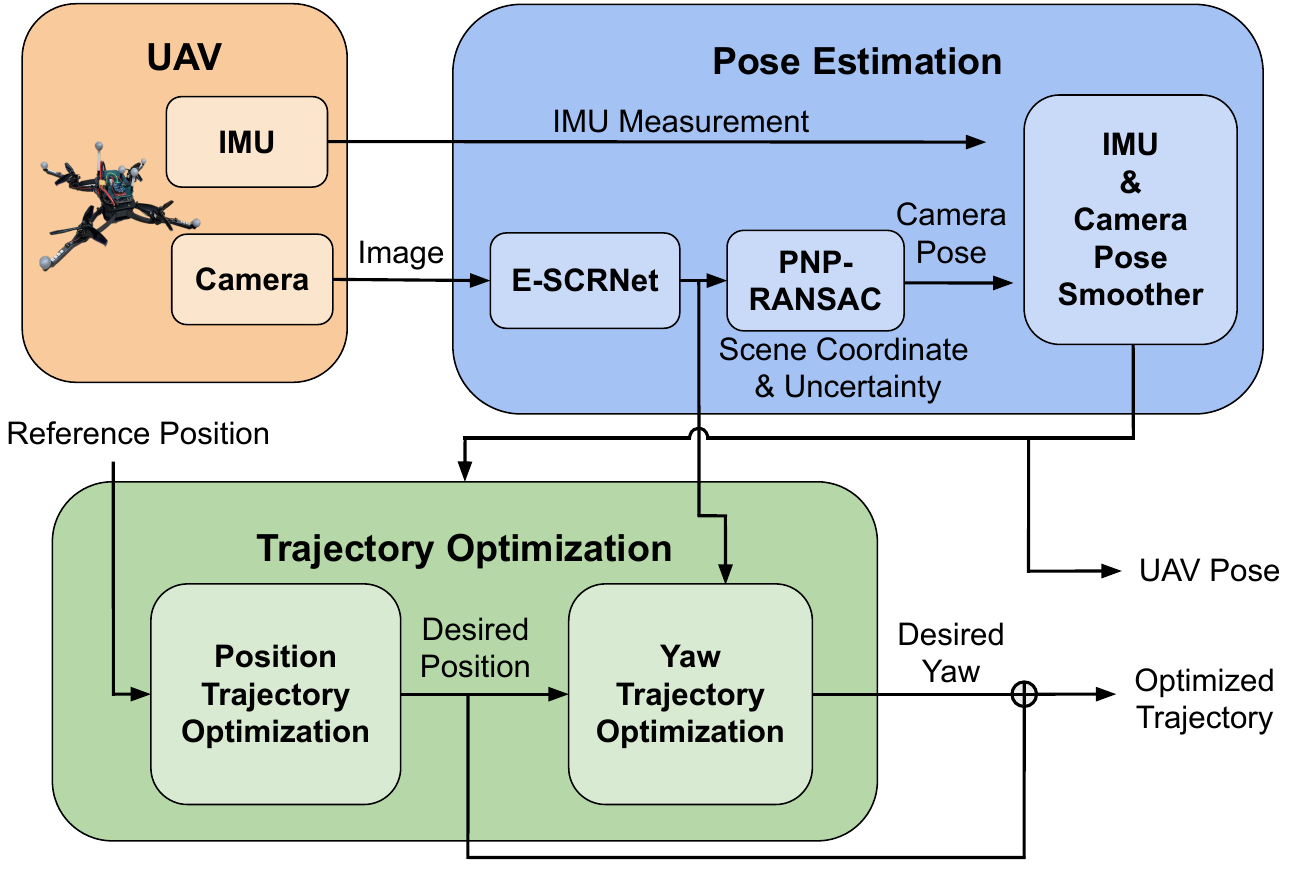}
		\caption{System architecture.}
		\label{fig:overview}
	\end{subfigure}
	\caption{Overview of the proposed approach. (a)~Camera orientations \textcolor{blue}{(blue)} are optimized to steer the camera toward low-entropy coordinates along the position trajectory \textcolor{red}{(red)}, starting from the current pose (black). (b)~The architecture comprises (1) a pose estimation block consisting of E-SCRNet, PnP-RANSAC, and a fixed-lag smoother, and (2) a two-stage trajectory optimization block for position and yaw.}
	\label{fig:overview_combined}
	\vspace{-0.25in}
\end{figure}

Scene coordinate regression (SCR) offers a learning-based alternative that predicts a 3D coordinate per pixel from a single image~\cite{Shotton13CVPR, DSAC17CVPR, DSACplusplus22TPAMI, Li20CVPR, Wang2023ArXiv, Brachmann23CVPR, Wang24CVPR, Jiang25CVPR}.
A Perspective-n-Point (PnP) solver then produces a drift-free absolute camera pose with high metric accuracy even over long trajectories, requiring only a compact neural network.
However, not all pixels yield reliable coordinates, and SCR often produces outliers and runs at low frequency with high latency, complicating its use in real-time planning and control.

This paper proposes a perception-aware trajectory planning algorithm, illustrated in Fig.~\ref{fig:hook}, that explicitly utilizes the pixel-wise uncertainty of an evidential learning-based SCR network (E-SCRNet).
To the best of our knowledge, this is the first active-perception work using SCR.
We train a neural network for SCR with evidential learning~\cite{Amini2020NeuRIPS}, so each pixel outputs scene coordinates and both aleatoric and epistemic uncertainties, which we summarize as entropy.
The entropy map drives a receding-horizon trajectory optimizer that (i) weighs low-entropy pixels higher and (ii) removes high-entropy pixels from the cost, thereby orienting the camera toward informative viewpoints while following the given reference position trajectory.
A fixed-lag smoother then fuses the low-rate SCR pose with high-rate IMU measurements, reducing latency, mitigating outlier SCR pose estimates, and eliminating drift caused by IMU bias.

The main contributions of this work are as follows:
\begin{itemize}
\item We present a real-time, receding-horizon trajectory planner that injects E-SCRNet uncertainty into perception-aware trajectory optimization, facilitating high-accuracy localization.
	\item We empirically show that entropy is a more informative uncertainty metric than separate aleatoric and epistemic uncertainty measures.
	\item We validate our approach through simulation and hardware-in-the-loop experiments, where the proposed planner outperforms baselines in localization.
\end{itemize}

%% file: section/related_work.tex
\subsection{Perception-Aware Trajectory and Path Planning}

PAMPC~\cite{PAMPC18IROS} augments nonlinear model predictive control (MPC) with a visibility cost, and Zhang et al.~\cite{Zhang18ICRA} frame perception-aware planning in a receding-horizon fashion.
Yaw optimization was later coupled with differential flatness and a differentiable camera cost~\cite{Murali19ACC}, extended to incorporate semantic priors~\cite{Bartolomei20IROS}, and parameterized time-optimally~\cite{Spasojevic20ICRA}.
Other directions incorporate information-theoretic~\cite{Zhang2020Arxiv} or topological metrics~\cite{Kim21IROS}.

More recent work combines position-yaw optimization with a novel feature-visibility model~\cite{APACE2024ICRA}, considers feature-limited or unknown environments~\cite{Yu2025ArXiv}, and introduces control barrier functions for safety guarantees~\cite{Kim25ICRA}, while other work is more data-driven~\cite{DiGiammarino24ECCV}.
However, all of these approaches rely on feature matching (FM) or visual-inertial odometry (VIO).
Unlike other methods, our work addresses perception-aware planning under scene coordinate regression (SCR).

\subsection{Learning-based Visual Localization}
Beyond FM and VIO, numerous learning-based visual localization methods have emerged.
Absolute Pose Regressors (APRs) employ neural networks to directly regress camera pose~\cite{PoseNet15ICCV,Kendall17CVPR, Walch17ICCV}.
Uncertainty-aware variants~\cite{Kendall16ICRA,CoordiNet22WACV} also estimate uncertainty, yet APRs remain markedly less accurate than both model-based pipelines and other learning-based alternatives~\cite{Sattler19CVPR}.

HLoc~\cite{Sarlin19CVPR} retrieves candidate keyframes with global descriptors and refines the pose via local feature matching.
While accurate, its memory and computation grow with the number of keyframes and can exceed onboard resources as the map grows.

SCR regresses a dense 3-D scene-coordinate map and then estimates the pose with PnP-RANSAC.
The evolution of SCR began with random forests~\cite{Shotton13CVPR}, which DSAC later replaced with differentiable RANSAC and CNNs~\cite{DSAC17CVPR,DSACplusplus22TPAMI} for end-to-end training.
Without depth supervision, ACE~\cite{Brachmann23CVPR} enables SCR to be trained in a few minutes.
Recent advances include hierarchical SCR~\cite{Li20CVPR,Wang2023ArXiv} and pretrained encoders for co-visibility~\cite{Wang24CVPR, Jiang25CVPR} to handle large scenes, with Jiang et al.~\cite{Jiang25CVPR} additionally improving robustness.
Although SCR attains high absolute accuracy in a single forward pass, it can produce outliers, and its application in robotics remains underexplored.

\subsection{Uncertainty Quantification}
Bayesian neural networks (BNNs) estimate aleatoric and epistemic uncertainty via Monte-Carlo dropout~\cite{MCDropout16ICML}, deep ensembles~\cite{Ensemble2017NeuRIPS}, or more general Bayesian learning frameworks~\cite{Kendall17NIPS}.
Uncertainty-aware APR~\cite{Kendall16ICRA} and VIO using BNNs and
neural radiance fields~\cite{NVINS24IROS} show the benefit of incorporating BNN uncertainty estimates for localization.
However, BNNs require multiple forward passes for uncertainty quantification.

On the other hand, evidential learning offers closed-form aleatoric and epistemic estimates without sampling~\cite{Amini2020NeuRIPS,Charpentier2022NatPN} and has been adopted for NeRF-aided SCR~\cite{Chen24ICRA}, semantic segmentation~\cite{Ancha24ICRA}, semantic mapping~\cite{Kim24IROS}, and traversability learning~\cite{EVORA24TRO}.
By applying evidential learning to our approach, the uncertainty of each predicted scene coordinate is quantified with only a single forward pass.

%% file: section/prob_and_overview.tex
We aim to improve localization along a given reference path by optimizing the vehicle’s position and yaw trajectories \emph{sequentially} rather than jointly: the position trajectory is solved first to track the reference, and the yaw trajectory is then solved to orient the camera.
Notably, the yaw is not constrained to the trajectory tangent or velocity direction; it is a decoupled degree of freedom dedicated to localization accuracy.

\textbf{Problem.}
At each step, we aim to compute a path of length $T_{\mathrm{plan}}$ in a receding-horizon fashion based on the following inputs:
(i) a reference trajectory, provided as an externally specified sequence of position waypoints with associated arrival times,
(ii) predicted scene coordinates and their associated uncertainties from E-SCRNet, and (iii) the current UAV pose estimate.
The objective is to produce a dynamically feasible trajectory that \emph{(a)} follows the reference, \emph{(b)} respects velocity/acceleration
limits, and \emph{(c)} orients viewpoints with low uncertainty to maximize localization accuracy.
Thanks to the differential flatness of multirotor dynamics~\cite{Mellinger11ICRA}, the position and yaw trajectories $\{\mathbf{p}(t),\psi(t)\}$ and their derivatives fully determine the required state and control.

\textbf{Pipeline.}
Fig.~\ref{fig:overview} depicts the pipeline for our receding-horizon trajectory generation framework.
Images from a camera pass through E-SCRNet~\cite{Chen24ICRA} to produce scene coordinates and uncertainties, which a PnP-RANSAC module converts into camera poses.
To compensate for the latency and outliers of this stream, we fuse it with high-rate IMU measurements using a fixed-lag smoother, yielding a clean, high-frequency state estimate.

This pose estimate, the most recent scene coordinates and uncertainties from the previous execution period $T_{\text{exec}}$, and reference waypoints feed a perception-aware trajectory optimization. This optimization is split into position and yaw subproblems that are solved sequentially in order to achieve the runtime efficiency required for real-time execution.
The resulting reference trajectory $\{\mathbf{p}^{\star}(t),\psi^{\star}(t)\}$ is continuously generated until the UAV reaches its goal pose.

%% file: section/method.tex
We first describe the training of E-SCRNet, an SCR network trained via deep evidential regression~\cite{Amini2020NeuRIPS}, and its uncertainty representation, then present our B-spline-based receding-horizon perception-aware trajectory planner.

\subsection{Scene Coordinate Regression with Deep Evidential Regression}

We train a neural network for SCR and assess the uncertainty of each predicted scene coordinate.
We adopt deep evidential regression~\cite{Amini2020NeuRIPS}, following the approach in~\cite{Chen24ICRA}, which jointly quantifies aleatoric and epistemic uncertainty in a single forward pass.
Unlike previous works~\cite{Amini2020NeuRIPS, Chen24ICRA}, our approach represents the total uncertainty as entropy to identify reliable scene coordinates.

Let $\mathbf{v} \in \mathbb{R}^3$ denote the scene coordinate of each pixel. For brevity, we write $v$ when referring to one Cartesian component.
We model $v \sim \mathcal{N}(\mu,\sigma^2)$, where the unknown mean and variance
$\mathbf{t}=(\mu,\sigma^2)$ are themselves random variables drawn from a
Gaussian and inverse gamma prior
$\mu \sim \mathcal{N}(\gamma,\sigma^2/\lambda\bigr)$, $\sigma^2 \sim \Gamma^{-1}(\alpha,\beta)$.
Assuming that the priors on $\mu$ and $\sigma^2$ are factorized, $p(\mathbf{t} \vert \mathbf{m})$ is the Normal Inverse-Gamma (NIG) distribution, which is a Gaussian conjugate prior parameterized by $\mathbf{m}=(\gamma,\lambda,\alpha,\beta)$ with $\lambda >0$, $\alpha > 1$, and $\beta > 0$.

The NIG distribution serves as a higher-order evidential distribution from which the lower-order Gaussian likelihood that generates scene coordinates is sampled, allowing the network to express the prediction ${\mathbb{E}[\mu] = \gamma}$, the aleatoric uncertainty $ \mathbb{E}[\sigma^2] = \beta/(\alpha - 1)$, and epistemic uncertainty $\text{Var}[\mu] = \beta/\big(\lambda(\alpha - 1)\big)$.

Deep evidential regression aims to learn the hyperparameters $\mathbf{m}_{\theta} = (\gamma_{\theta}, \lambda_{\theta}, \alpha_{\theta}, \beta_{\theta})$. The predictive distribution of the scene coordinate is obtained by marginalizing the Gaussian likelihood over the latent $\mathbf{t}=(\mu,\sigma^2)$ with respect to the NIG prior, $p(v \vert \mathbf{m}_{\theta}) = \int p(v \vert \mathbf{t})\,p(\mathbf{t} \vert \mathbf{m}_{\theta})\,d\mathbf{t}$. Because the NIG is the conjugate prior of the Gaussian, this integral admits the closed-form Student-t
$p(v \vert \mathbf{m}_{\theta}) = \text{St}\big(v ; \gamma_{\theta}, \beta_{\theta}(1+\lambda_{\theta})/(\lambda_{\theta}\alpha_{\theta}) , 2\alpha_{\theta} \big),
$
where $\text{St}\big(v ; \eta, \tau^2, \chi)$ is a Student-t distribution with location $\eta$, scale $\tau^2$, and degrees of freedom $\chi$.

The total loss~$\mathcal{L}_{\theta}$ minimized during training is the mean of the per-sample loss
${\mathcal{L}_{\theta}^i = \mathcal{L}_{\theta,\mathrm{nll}}^i + 
\rho\,\mathcal{L}_{\theta,\mathrm{reg}}^i}$.
The loss $\mathcal{L}_{\theta, \mathrm{nll}}^i$ is chosen as the negative log-likelihood~(NLL) of $p(v \vert \mathbf{m}_{\theta})$ for maximum likelihood estimation~(MLE):
\begin{equation} \label{eq:nll_loss}
\begin{split}
&\mathcal{L}_{\theta, \mathrm{nll}}^i =
\log\!\left(
\Gamma(\alpha_{\theta}) / \Gamma\!\left(\alpha_{\theta} + \tfrac12\right)
\right)
\\
&+ \tfrac12 \log\!\left( \pi / \lambda_{\theta} \right)
- \alpha_{\theta} \log\!\left(2\beta_{\theta} (1+\lambda_{\theta})\right) \\
&+ 
\bigl(\alpha_{\theta} + \tfrac12\bigr)
\log\!\left(\lambda_{\theta}(v_i-\gamma_{\theta})^2 + 2\beta_{\theta}(1+\lambda_{\theta})\right).
\end{split}
\end{equation}

The loss $\mathcal{L}_{\theta, \mathrm{reg}}$ is designed for regularization:
\begin{equation}  \label{eq:reg_loss}
\mathcal{L}_{\theta, \mathrm{reg}}^i = (2\lambda_{\theta} + \alpha_{\theta})\vert v_i - \gamma_{\theta} \vert.
\end{equation}

After training, we can express the predictive entropy $H_{\theta} = \mathbb{E}_{v \sim p(v \vert \mathbf{m}_{\theta})}[ -\log p(v \vert \mathbf{m}_{\theta})]$ in its closed form
\begin{equation} \label{eq:entropy}
\begin{split}
H_{\theta} &=\!(\alpha_{\theta} + \tfrac12 ) ( \psi(\alpha_{\theta} + \tfrac12) - \psi(\alpha_{\theta}) ) \\
&+\tfrac12\log\!\left( 2\pi\beta_{\theta}(1 + \lambda_{\theta}) / \lambda_{\theta} \right) \\
&+ \log\!\left(
\Gamma(\alpha_{\theta}) / \Gamma\!\left(\alpha_{\theta} + \tfrac12\right)
\right),
\end{split}
\end{equation}
where $\psi(\cdot)$ is the digamma function.
We use entropy as the uncertainty representation because it captures both aleatoric and epistemic uncertainty.
The detailed justification is given in Section~\ref{subsec:uncertainty}.
We refer the reader to~\cite{Amini2020NeuRIPS} for more information on deep evidential regression.

\subsection{Trajectory Optimization}
We receive the current UAV pose as estimated by the fixed-lag smoother alongside reference waypoints and scene coordinates with their corresponding uncertainties.
\rev{Given these inputs, position and yaw trajectories are optimized to track waypoints and orient the camera toward reliable scene coordinates.}
In this section, we detail the perception-aware trajectory optimization.

\textbf{Uniform B-Spline.} Both position and yaw are parameterized as clamped, uniform B-splines of degree $k=3$:
\begin{equation} \label{eq:bspline}
	\mathbf{q}(t) = \sum_{i=0}^{n}\mathbf{q}_iN_{i,k}(t)
\end{equation}
where the knot vector is $\{t_{0},t_{1},\dots,t_{n+k+1}\}$, the control points are
$\mathbf{q}_{i}=\{q_{x,i},q_{y,i},q_{z,i},q_{\psi,i}\}\in\mathbb{R}^{4}$,
and $N_{i,k}(t)$ denote the B-spline basis functions.
Each knot span has equal duration $\Delta t=t_{i+1}-t_{i}$.  
For clamping, the first and last $k$ knots are repeated:
$t_{0}=\dots=t_{k-1}$ and $t_{n+1}=\dots=t_{n+k+1}$.
We fix the first control point to the current UAV pose and the last control point to the final waypoint within the planning horizon~$T_{\mathrm{plan}}$.
The other control points become decision variables during the optimization.

\textbf{Optimization Cost.} Following Bartolomei et al.~\cite{Bartolomei20IROS}, we represent the total cost function $C_{\mathrm{total}}$ as a weighted sum of cost terms serving different purposes.
To reduce computational load, we optimize the position and yaw sequentially by minimizing two disjoint sub-objectives rather than the full $C_{\mathrm{total}}$:
\begin{equation} \label{eq:disjoint_cost}
\begin{gathered}
C_{\mathrm{total}}^{\mathbf{p}} = \lambda_{\mathrm{wp}} C_{\mathrm{wp}}^{\mathbf{p}} + \lambda_{\mathrm{eq}} C_{\mathrm{eq}}^{\mathbf{p}} + \lambda_{\mathrm{ie}} C_{\mathrm{ie}}^{\mathbf{p}} + \lambda_s C_{s}^{\mathbf{p}} \\
C_{\mathrm{total}}^{\psi} = \lambda_{\mathrm{fov}} C_{\mathrm{fov}}^{\psi} + \lambda_{\mathrm{eq}} C_{\mathrm{eq}}^{\psi} + \lambda_{\mathrm{ie}} C_{\mathrm{ie}}^{\psi} + \lambda_s C_{s}^{\psi}.
\end{gathered}
\end{equation}
where the superscripts, $\mathbf{p}$ and $\psi$, indicate that the terms are evaluated on the position and yaw trajectory, respectively.
For brevity, we will drop the superscripts in the following.
The coefficients $\lambda_{(\cdot)}$ balance five components: 
$C_{\mathrm{wp}}$ penalizes deviation from the given waypoints, 
$C_{\mathrm{fov}}$ rewards keeping reliable scene coordinates within the camera’s field-of-view (FOV), 
$C_{\mathrm{eq}}$ enforces the trajectory starts from the given initial velocity, acceleration, and jerk, 
$C_{\mathrm{ie}}$ enforces the specified velocity and acceleration bounds, 
and $C_{s}$ promotes smoothness of the trajectory.

Let $\{\mathbf{w}_{0},\dots,\mathbf{w}_{r-1}\}$ denote all waypoints and let $\{s_{1},\dots,s_{r-2}\}$ be the prescribed arrival times for all but the first and last.  
The waypoint cost $C_{\mathrm{wp}}$ penalizes the squared Euclidean error between the position trajectory and each desired waypoint:
\begin{equation} \label{eq:waypoint_cost}
C_{\mathrm{wp}}
= \sum_{i=1}^{r-2} \,\sum_{o\in\{x,y,z\}}
\bigl(q_{o}(s_{i}) - w_{o,i}\bigr)^{2}.
\end{equation}

Our formulation of the FOV cost $C_{\mathrm{fov}}$ is inspired by Murali et al.~\cite{Murali19ACC}.
Whereas the original cost encourages the camera to track as many covisible features as possible within its frustum, our version instead rewards keeping low-uncertainty scene coordinates in view regardless of covisibility.

We assume the camera is mounted on the UAV with a known relative pose.
Given the trajectories, we can compute the camera position $\mathbf{p}_{C,i}\in\mathbb{R}^{3}$ and orientation $R_{i}^{C}\in\mathrm{SO}(3)$ with respect to the world frame at each time $t=s_{i}$.
By differential flatness~\cite{Mellinger11ICRA}, both are determined by the position and yaw trajectories,
\begin{equation} \label{eq:cam_pose}
\mathbf{p}_{C,i}=\mathbf{q}_{p}(s_{i}),
\qquad
R_{i}^{C}=R_{B}^{C}\,R_{i}^{B},
\end{equation}
where $\mathbf{q}_{p}(t)=[q_{x}(t)\;\;q_{y}(t)\;\;q_{z}(t)]^{\top}$ is the position trajectory, $R_{B}^{C}$ is the known body-to-camera extrinsic rotation, and the world-to-body rotation $R_{i}^{B}=[\mathbf{x}_{b}\;\;\mathbf{y}_{b}\;\;\mathbf{z}_{b}]^{\top}$ is recovered from the flat outputs.

Next, we build a differentiable indicator function describing whether scene coordinates $\mathbf{v}^C_{i,j}$ are within the camera frustum.
For this purpose, we consider the displacement vectors from the optical center to each of the four image-plane corners:
\begin{equation} \label{eq:vertices}
\begin{split}
\mathbf{z}_{TR} &= \big[(w - c_x)l_x \quad (- c_y)l_y \quad f \big]^{\top} \\
\mathbf{z}_{LR} &= \big[(w - c_x)l_x \quad (h - c_y)l_y \quad f \big]^{\top} \\
\mathbf{z}_{TL} &= \big[(- c_x)l_x \quad (- c_y)l_y \quad f \big]^{\top} \\
\mathbf{z}_{LL} &= \big[(- c_x)l_x \quad (h - c_y)l_y \quad f \big]^{\top}.
\end{split}
\end{equation}
Here, $w$ and $h$ are the image width and height in pixels, $(c_{x},c_{y})$ is the principal point in pixels, $l_{x}$ and $l_{y}$ are the physical pixel sizes $(m/\text{pix})$, and $f$ is the focal length in meters.
The normal vectors of the four side half planes of the camera frustum are
$\mathbf{z}_{\mathrm{TR}}\!\times\!\mathbf{z}_{\mathrm{LR}}$,
$\mathbf{z}_{\mathrm{TL}}\!\times\!\mathbf{z}_{\mathrm{TR}}$,
$\mathbf{z}_{\mathrm{LL}}\!\times\!\mathbf{z}_{\mathrm{TL}}$, and
$\mathbf{z}_{\mathrm{LR}}\!\times\!\mathbf{z}_{\mathrm{LL}}$.
Using these normals, we construct a soft indicator vector
\begin{equation} \label{eq:indicator_vec}
\mathbf{o}_{i,j} =
\frac12
\begin{bmatrix}
1+\tanh\!\left((\mathbf{z}_{\mathrm{TR}}\!\times\!\mathbf{z}_{\mathrm{LR}})\!\cdot\!\mathbf{v}^{C}_{i,j}/s\right) \\[2pt]
1+\tanh\!\left((\mathbf{z}_{\mathrm{TL}}\!\times\!\mathbf{z}_{\mathrm{TR}})\!\cdot\!\mathbf{v}^{C}_{i,j}/s\right) \\[2pt]
1+\tanh\!\left((\mathbf{z}_{\mathrm{LL}}\!\times\!\mathbf{z}_{\mathrm{TL}})\!\cdot\!\mathbf{v}^{C}_{i,j}/s\right) \\[2pt]
1+\tanh\!\left((\mathbf{z}_{\mathrm{LR}}\!\times\!\mathbf{z}_{\mathrm{LL}})\!\cdot\!\mathbf{v}^{C}_{i,j}/s\right) \\[2pt]
1+\tanh\!\left((\mathbf{v}^{C}_{i,j}-f\mathbf{e}_{3})\!\cdot\!\mathbf{e}_{3}/s\right)
\end{bmatrix},
\end{equation}
where $\mathbf{e}_{3}=[0,0,1]^{\top}$ and $s>0$ are the camera’s forward axis and a smoothing constant, respectively. 
For the $j$-th scene coordinate at time $t=s_{i}$, the desired scalar indicator is defined as the product of the elements of $\mathbf{o}_{i,j}$:
\begin{equation} \label{eq:indicator_function}
F(\mathbf{c}_i, \mathbf{v}_j) =  \prod_{k=1}^5 o_{i,j,k}.
\end{equation}
The scalar indicator approaches 1 when the point lies inside the frustum and 0 otherwise.  
A larger $s$ yields a smoother transition but a less selective indicator.

The cost is the weighted sum of the indicators over all sampling times and scene coordinates:
\begin{equation} \label{eq:fov_cost}
C_{fov} = -\sum_{i=1}^{r-1}\sum_{j=1}^{n_f} \exp(-a_{\mathrm{fov}} H_{\theta,j}) F(\mathbf{c}_i, \mathbf{v}_j).
\end{equation}
Here, $n_f$ denotes the number of selected scene coordinates.
These coordinates are randomly chosen among those with entropy values below a predefined threshold.
The weight $\exp(-a_{\mathrm{fov}} H_{\theta,j})$ depends on the entropy \(H_{\theta,j}\) from Eq.~\eqref{eq:entropy}; $a_{\mathrm{fov}}>0$ is constant.  
Lower entropy, i.e., higher certainty, yields a larger weight, encouraging the yaw trajectory to keep such scene coordinates in view.

Because we solve the optimization repeatedly over a moving horizon \(T_{\text{exec}}\), each new trajectory must splice smoothly onto the previous one.  
To enforce continuity, the initial-equality cost $C_{\text{eq}}$ penalizes any mismatch between the prescribed initial derivatives and those of the candidate trajectory at $t=t_0$:
\begin{equation} \label{eq:equality_cost}
C_{\text{eq}}
= \Vert \mathbf{v}(t_0)-\mathbf{v}_{0}\Vert_{2}^{2}
+ \Vert \mathbf{a}(t_0)-\mathbf{a}_{0}\Vert_{2}^{2}
+ \Vert \mathbf{j}(t_0)-\mathbf{j}_{0}\Vert_{2}^{2}
\end{equation}
where \(\mathbf{v}_{0},\mathbf{a}_{0},\mathbf{j}_{0}\in\mathbb{R}^{4}\) denote the desired initial velocity, acceleration, and jerk for the translation coordinates and yaw.
Note that the initial position and yaw automatically match their desired values because a clamped B-spline is anchored to the first control point.

The inequality cost $C_{\text{ie}}$ enforces that every component of the planned trajectory respects user-specified velocity and acceleration limits:
\begin{equation} \label{eq:ineq_cost}
C_{\text{ie}}=\!\!\sum_{i=1}^{r-1}\!\sum_{o\in\{x,y,z,\psi\}}\!\bigl(
M_{v}(v_{o}(s_{i})) + M_{a}(a_{o}(s_{i}))\bigr),
\end{equation}
where $v_{o}(s_{i})$ and $a_{o}(s_{i})$ denote, respectively, the velocity and acceleration of coordinate \(o\) at sample time \(s_{i}\).
The function $M_u(u_o)=\max\big((u_o^2-u_{\max}^2)^{2}, 0 \big)$ is applied to both velocity $M_v(v_o)$ and acceleration $M_a(a_o)$.

The smoothness cost $C_{\text{s}}$ promotes a smooth trajectory by enforcing consistency between consecutive elastic bands, $\mathbf{S}_{i+1,i}$ and $\mathbf{S}_{i-1,i}$ with $\mathbf{S}_{i,j} = \mathbf{q}_{i}-\mathbf{q}_{j}$~\cite{Zhu15CDC}:
\begin{equation}
C_{\text{s}}
= \sum_{i=1}^{n-1}
\bigl\lVert \mathbf{S}_{i+1,i} + \mathbf{S}_{i-1,i} \bigr\rVert_{2}^{2}.
\end{equation}

%% file: section/experiments.tex
\begin{figure}[tb!]
	\centering
	\includegraphics[width=0.7\linewidth]{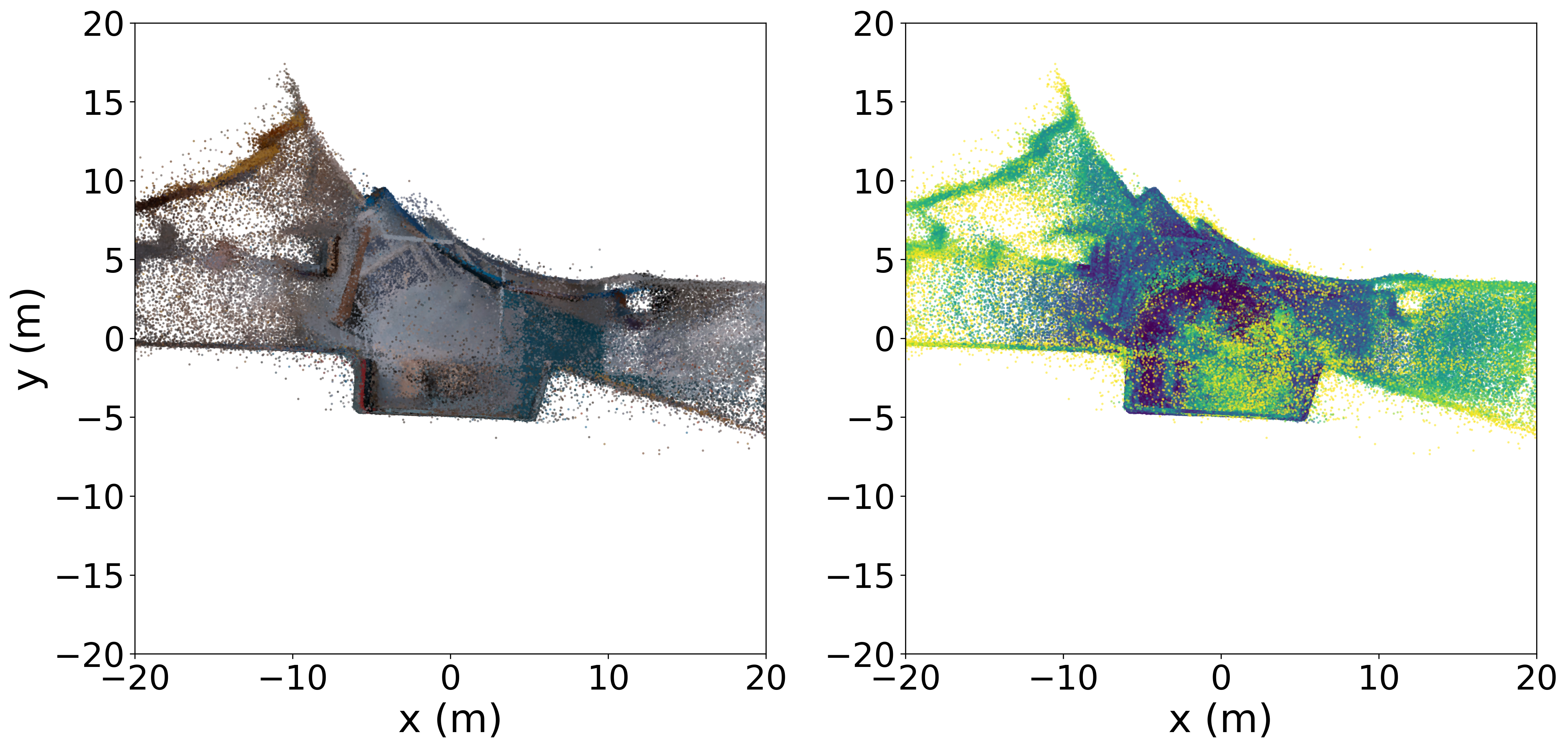}
	\caption{Top-down 3-D reconstruction of the \texttt{stata1} site produced by SCR. (left) RGB-colored points (right) Normalized entropy map. The accumulated scene coordinates and their uncertainties from multiple viewpoints reveal spatially consistent uncertainty distributions across the environment.}
	\label{fig:map}
	\vspace{-0.25in}
\end{figure}

We begin by evaluating our approach in simulation.
Our evaluation is based on two key assumptions: (A1) lower uncertainty implies higher scene-coordinate accuracy, which we validate empirically in Section~\ref{subsec:uncertainty}, and (A2) uncertainty varies smoothly over the trajectory because it is spatially correlated. This spatial correlation allows the uncertainty to be used within short planning horizons, as supported by Fig.~\ref{fig:map} and the supplementary video: \href{https://www.youtube.com/watch?v=uYPff9E5mQU}{\textcolor{linkblue}{\texttt{youtube.com/watch?v=uYPff9E5mQU}}}.

Our two goals are:
(a) to verify that the entropy defined in Eq.~\eqref{eq:entropy} provides an appropriate measure of uncertainty, by comparing it with conventional aleatoric and epistemic uncertainty measures, and
(b) to demonstrate the effectiveness of our perception-aware trajectory planner against baselines.
We also conduct hardware-in-the-loop experiments to validate system-level integration with onboard IMU data and simulated visual input.

\begin{figure}[t]
	\centering
	\begin{subfigure}[b]{0.4\linewidth}
		\centering
		\includegraphics[width=\linewidth]{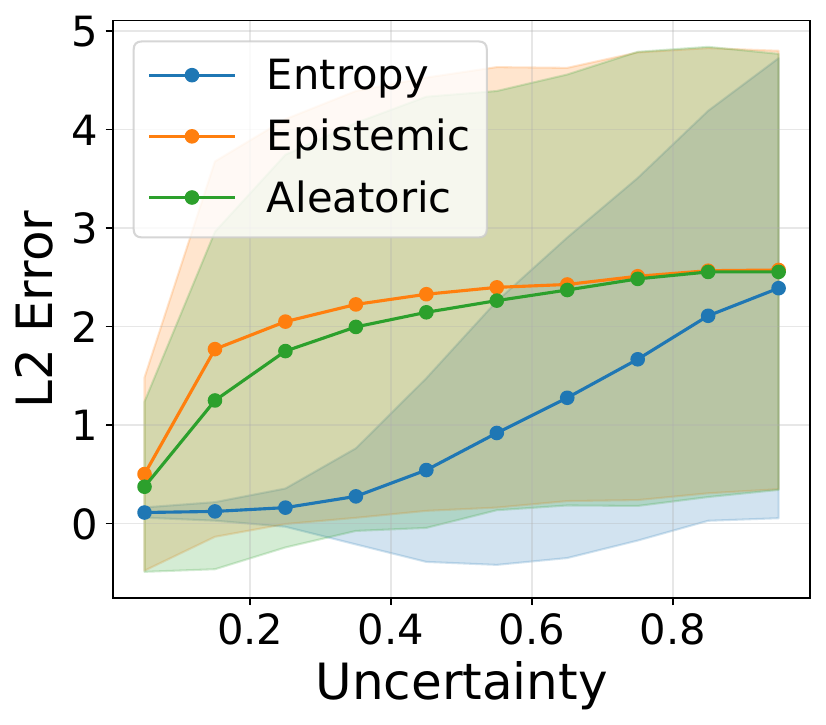}
		\label{fig:l2error_v_unc}
	\end{subfigure}\hspace{2pt}
	\begin{subfigure}[b]{0.4\linewidth}
		\centering
		\includegraphics[width=\linewidth]{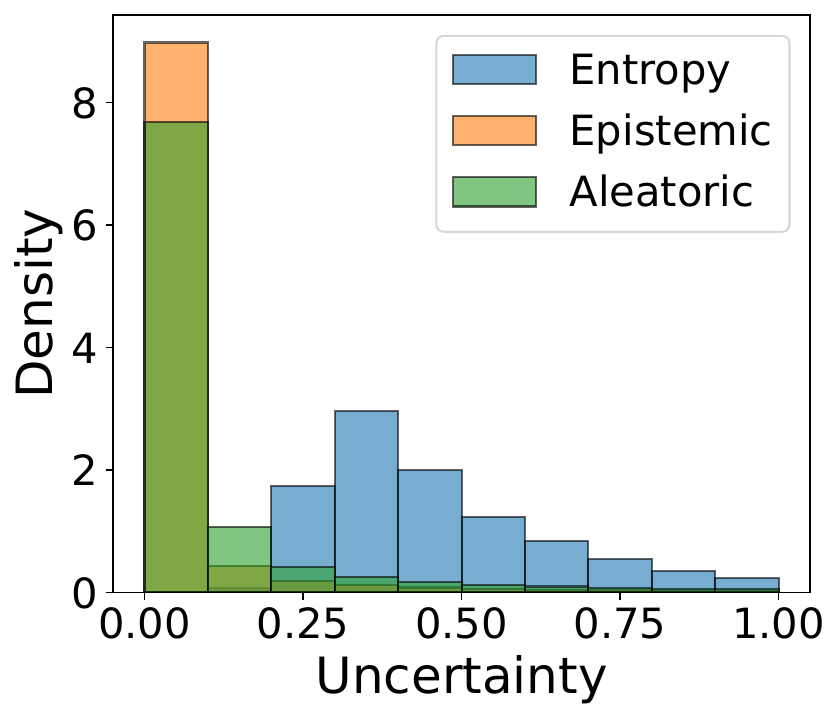}
		\label{fig:hist_v_unc}
	\end{subfigure}
	\vspace{-0.15in}
	\caption{Plots showing normalized uncertainty and scene coordinate error of E-SCRNet predictions. (left) Mean and standard deviation of L2 error (right) Density histogram.}
	\label{fig:graph_uncertainty}
	\vspace{-0.1in}
\end{figure}

\begin{figure}[tb!]
	\centering
	\begin{subfigure}[htb]{0.25\linewidth}
		\centering
		\includegraphics[width=\linewidth]{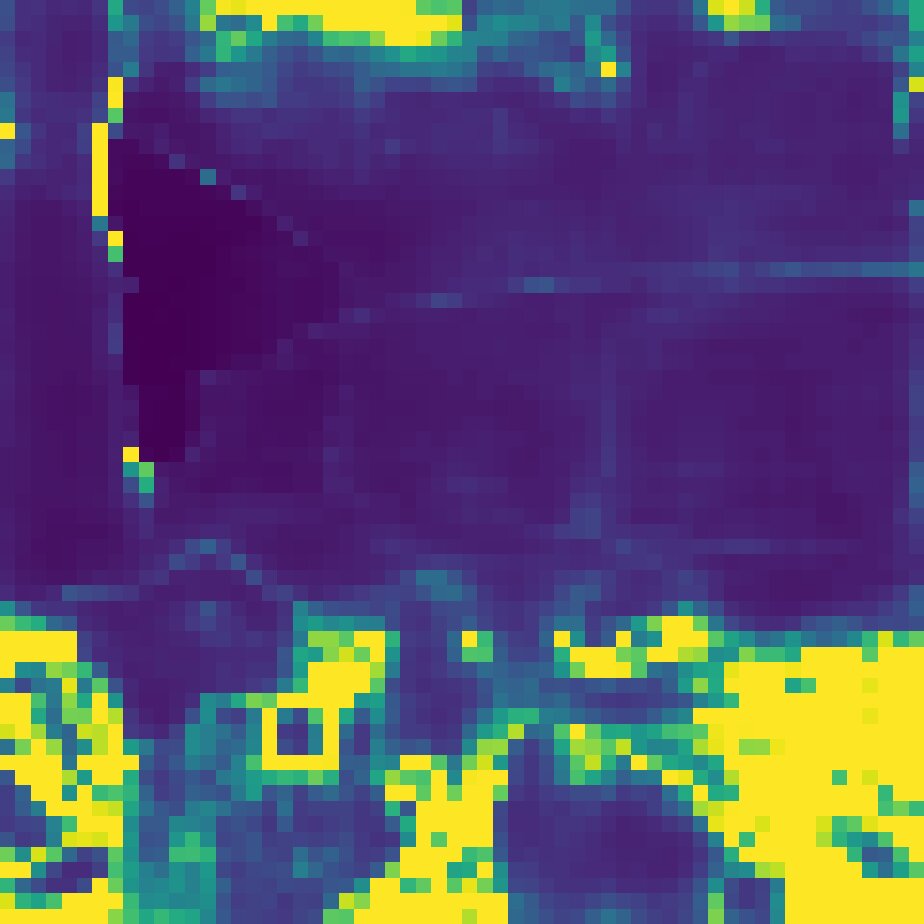}
		\label{fig:img1}
	\end{subfigure}\hspace{2pt}
	\begin{subfigure}[htb]{0.25\linewidth}
		\centering
		\includegraphics[width=\linewidth]{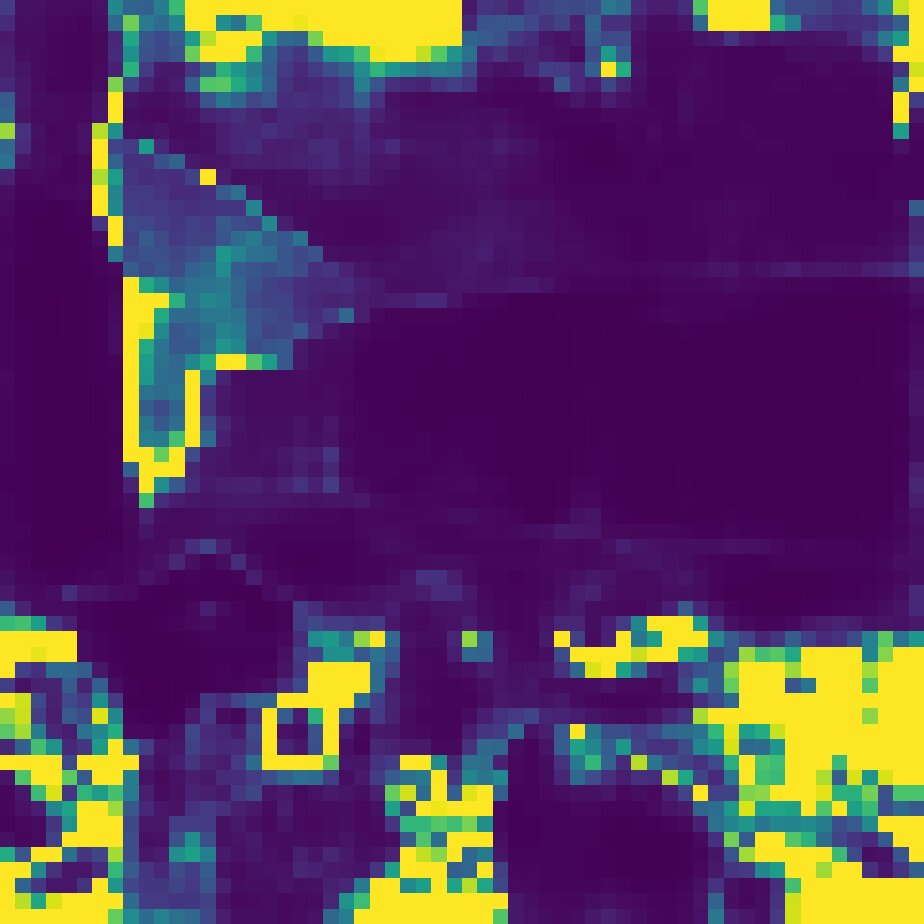}
		\label{fig:img2}
	\end{subfigure}\hspace{2pt}
	\begin{subfigure}[htb]{0.25\linewidth}
		\centering
		\includegraphics[width=\linewidth]{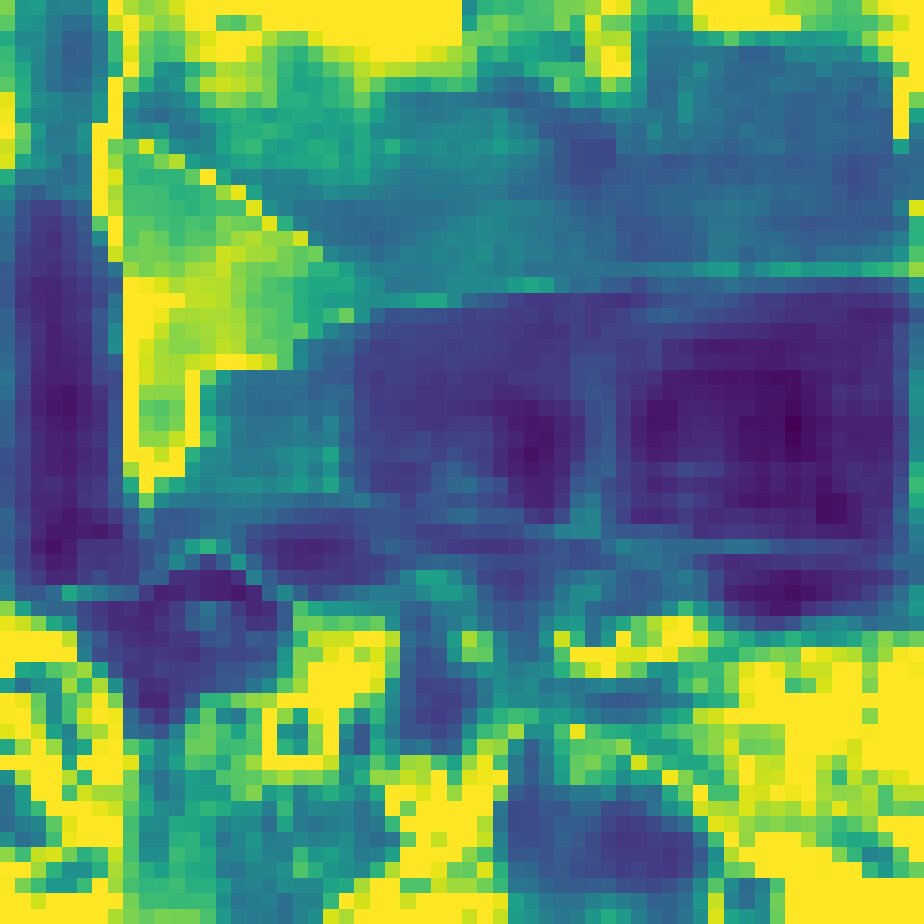}
		\label{fig:img3}
	\end{subfigure}\\[1mm]
	\begin{subfigure}[htb]{0.25\linewidth}
		\centering
		\includegraphics[width=\linewidth]{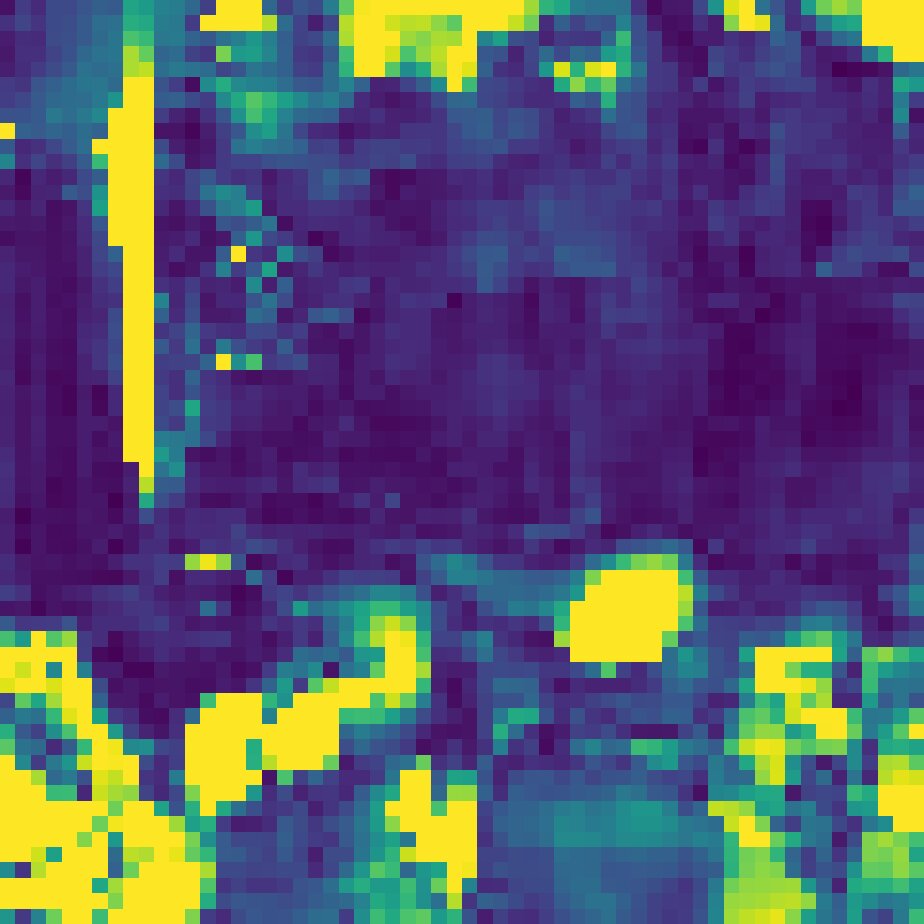}
		\label{fig:img4}
	\end{subfigure}\hspace{2pt}
	\begin{subfigure}[htb]{0.25\linewidth}
		\centering
		\includegraphics[width=\linewidth]{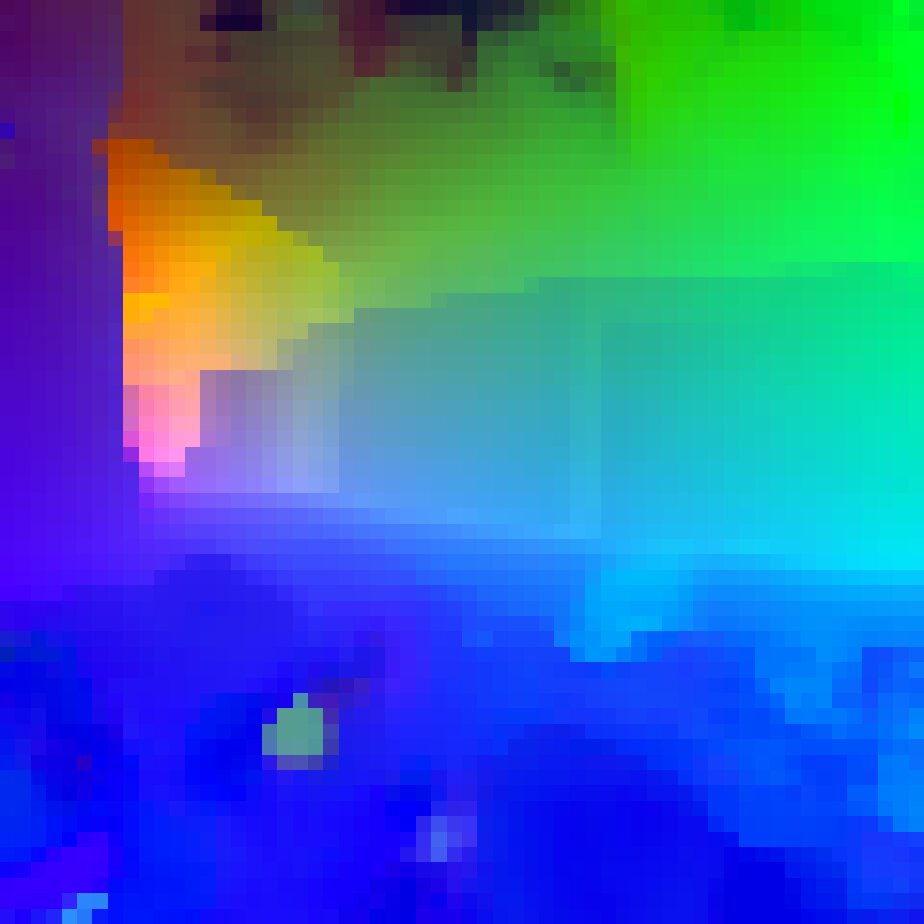}
		\label{fig:img5}
	\end{subfigure}\hspace{2pt}
	\begin{subfigure}[htb]{0.25\linewidth}
		\centering
		\includegraphics[width=\linewidth]{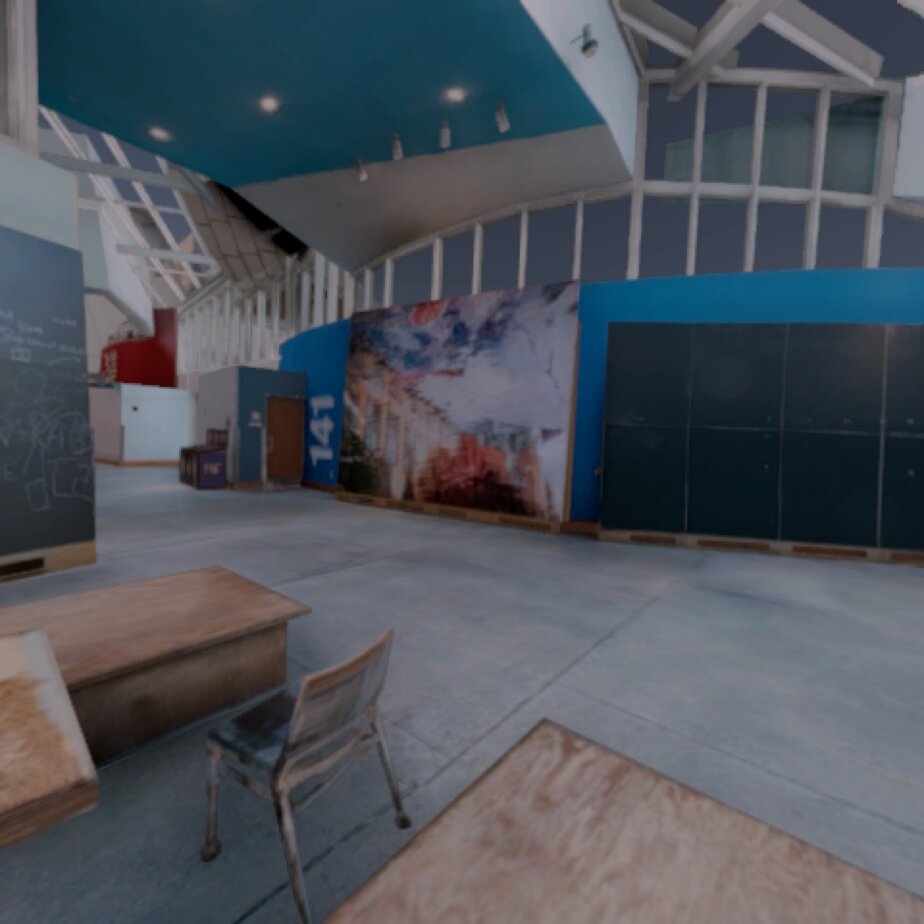}
		\label{fig:img6}
	\end{subfigure}
	\caption{Qualitative results of E-SCRNet for a test image. (upper left) Aleatoric Uncertainty (upper middle) Epistemic Uncertainty (upper right) Entropy (lower left) L2 Error of predicted scene coordinates (lower middle) Predicted scene coordinates (lower right) Input image.}
	\label{fig:qualitative}
	\vspace{-0.25in}
\end{figure}

\begin{figure*}[htb!]
	\centering         
	\begin{subfigure}[t]{\linewidth}
		\centering
		\includegraphics[width=.85\linewidth]{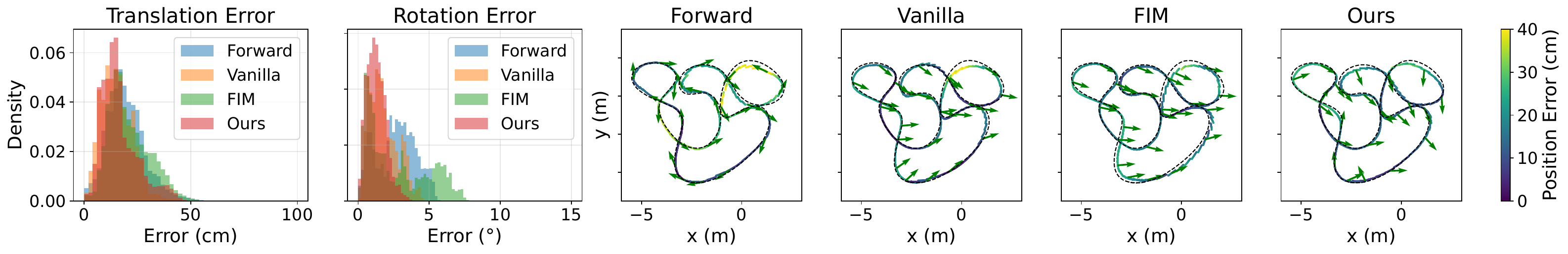}
		\caption{winter}\label{fig:winter_visualization}
	\end{subfigure}
	\begin{subfigure}[t]{\linewidth}
		\centering
		\includegraphics[width=.85\linewidth]{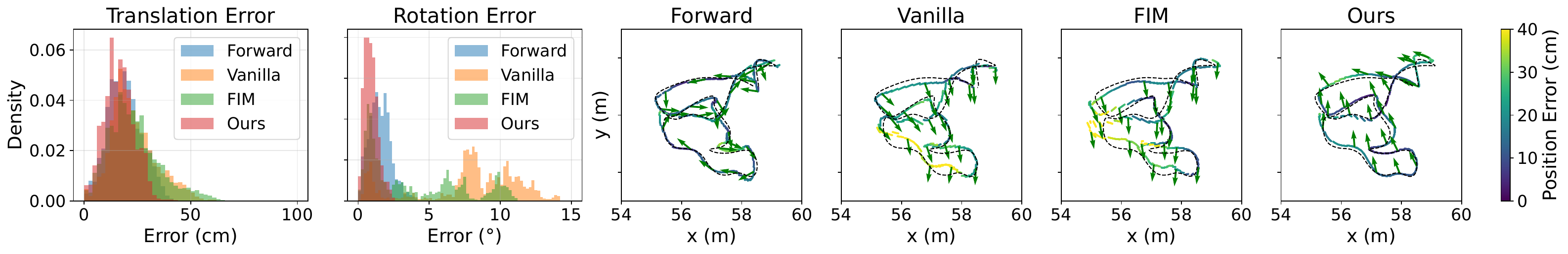}
		\caption{ampersand}\label{fig:ampersand_visualization}
	\end{subfigure}\hfill
	\caption{Comparison of baseline planners (\emph{Forward}, \emph{Vanilla}, \emph{FIM}) with the proposed method on ``winter'' and ``ampersand'' trajectories. (left) Density distributions of translation and rotation errors of the fixed-lag smoother for \emph{Forward} (blue), \emph{Vanilla} (orange), \emph{FIM} (green), and \emph{Ours} (red) over the full duration of each trajectory. (right) Top-down view showing reference trajectory (black dotted), the predicted trajectories (colored), camera orientations (green arrows) for one repetition of each trajectory.}
	\label{fig:visual_results}
	\vspace{-0.25in}
\end{figure*}

\subsection{Setup}
We run photorealistic simulations using the FlightGoggles simulator on an Intel Core i9-7920X CPU~\cite{flightgoggles}.
The simulated camera resolution is fixed at $480\times480$ pixels, with an FOV of $90^{\circ}$.
We train the E-SCRNet following the architecture in~\cite{DSACplusplus22TPAMI} for each of the two environments \texttt{stata1} and \texttt{stata2}.

To highlight the advantages of our perception-aware planner, we deliberately use a small training set captured along a single trajectory (``picasso'') from the Blackbird dataset~\cite{Antonini20IJRR} and train for 100k iterations.
Among the Blackbird trajectories, ``picasso'' is the most random, sweeping the full position bounds, and we keep the yaw continuously rotating so the network observes a wide range of viewing directions.
We set the regularization coefficient $\rho = 10^{-2}$.
Following~\cite{DSACplusplus22TPAMI}, we train the network using Adam with a learning rate of $10^{-4}$ and a batch size of 1, completing in 14 minutes on one NVIDIA RTX 6000 Ada GPU.

For every planning step, we minimize the costs in Eq.~\eqref{eq:disjoint_cost}
with the L-BFGS algorithm provided by NLopt~\cite{NLopt}.
The optimization parameters are ($\lambda_{\mathrm{wp}}$, $\lambda_{\mathrm{fov}}$, $\lambda_{\mathrm{eq}}$, $\lambda_{\mathrm{ie}}$, $\lambda_{s}$, $T_{\mathrm{plan}}$, $T_{\mathrm{exec}}$, $n_f$, $s$, $a_{\mathrm{fov}}$, $v_{\{x,y,z\},\mathrm{max}}$, $v_{\psi,\mathrm{max}}$, $a_{\{x,y,z\},\mathrm{max}}$, $a_{\psi,\mathrm{max}})$ $=$ $(10^{4}$, $10^{1}$, $10^{3}$, 1.0, 5.0, 0.8\,s, 0.5\,s, 200, 5.0, 0.5, 3.0\,m/s, 5.0\,rad/s, 20.0\,m/$\text{s}^2$, 25.0\,rad/$\text{s}^2$).
These values were tuned on a single circular trajectory, yet transfer to all eight evaluation trajectories across both environments without retuning, indicating low sensitivity to their exact setting.
The number of control points for the clamped B-spline in Eq.~\eqref{eq:bspline} is set to six.
Scene-coordinate regression is run every $0.2\,\mathrm{s}$.

For the hardware-in-the-loop experiments, the quadrotor flies within an arena while onboard IMU data is logged and the camera stream is simulated with images rendered from the virtual \texttt{stata1} scene, as shown in Fig.~\ref{fig:drone}.
The quadrotor's pose is tracked using a motion capture system.
An external machine equipped with an NVIDIA RTX A5000 GPU processes simulator images and executes SCR inference, smoothing, and optimization, transmitting control commands wirelessly to the UAV.

\subsection{Uncertainty Representation} \label{subsec:uncertainty}
We collect scene coordinate predictions and their uncertainty maps from E-SCRNet while the UAV follows a circular trajectory.
Fig.~\ref{fig:graph_uncertainty} shows the relationship between the L2 prediction error and three uncertainty estimates: aleatoric, epistemic, and entropy.
We clip the largest 3\% of uncertainty values and normalize the remainder to $[0,1]$ for visual clarity.

As shown in Fig.~\ref{fig:graph_uncertainty}, the mean error increases with increasing entropy, whereas its correlation with the others is weaker.
Moreover, the error's standard deviation declines as entropy decreases, while it remains high across the full range of the other measures.  
This validates assumption A1.
Fig.~\ref{fig:graph_uncertainty} further shows that entropy is less skewed toward zero than the other uncertainties, providing a more informative, well-spread metric, while Fig.~\ref{fig:qualitative} shows qualitative uncertainty and error maps for a representative test image.

\subsection{Simulation Results.} \label{subsec:simulation_result}

\textbf{Comparison to baselines.}
Previous VIO-based perception-aware planners~\cite{Murali19ACC, Bartolomei20IROS} use an FOV cost that steers the camera toward visual features to improve VIO accuracy, but employ no uncertainty metric.
In contrast, our method leverages E-SCRNet uncertainties to select reliable scene coordinates.
To demonstrate the effectiveness of this additional uncertainty information, we compare our planner with three yaw strategies: \emph{Forward}, which keeps the camera aligned with the velocity vector; \emph{Vanilla}, which uses the same perception-aware optimization but \emph{without E-SCRNet uncertainty}, i.e., a fixed weight of 1.0 in Eq.~\eqref{eq:fov_cost} and no high-entropy pixel filtering, matching previous strategies~\cite{Murali19ACC, Bartolomei20IROS} while still using SCR for fairness; and \emph{Fisher information matrix (FIM)}, which replaces the entropy weight in Eq.~\eqref{eq:fov_cost} with the trace of the FIM from Zhang et al.~\cite{Zhang2020Arxiv}.
We evaluate four trajectories from the Blackbird dataset per site, ``clover'', ``patrick'', ``thrice'', and ``winter'' in \texttt{stata1}, and ``ampersand'', ``dice'', ``halfmoon'', and ``sid'' in \texttt{stata2}.
Each trajectory is flown multiple times.

Tables~\ref{tab:comparison_imu} and~\ref{tab:comparison_scr} report localization accuracy for the fixed-lag smoother that fuses IMU and SCR (``\emph{IMU+SCR}'') and the raw SCR output through PnP-RANSAC (``\emph{SCR}''), respectively.
Our method consistently outperforms the baselines on ``\emph{IMU+SCR}'' localization,
reducing average of translation RMSE by 15.0\%, 4.9\%, and 23.5\% and that of rotation RMSE by 30.8\%, 38.6\%, and 41.3\% compared to \emph{Forward}, \emph{Vanilla}, and \emph{FIM}, respectively.
Average of mean translation errors are reduced by 9.4\%, 5.4\%, and 19.7\%, while that of mean rotation errors decrease by 27.3\%, 38.5\%, and 35.1\% relative to the baselines.
As the positional gap over the strongest baseline \emph{Vanilla} is small, we assess it with a block-level bootstrap ($100$ blocks/trajectory, $10{,}000$ replicates) that respects temporal correlation; the $95\%$ confidence interval for the mean improvement, $[0.27,\,1.84]\,$cm, excludes zero, confirming significance over \emph{Vanilla}.
For ``\emph{SCR}'' output, our method achieves the lowest mean errors compared to all baselines.
Fig.~\ref{fig:visual_results} shows estimated positions, camera orientations, and error histograms for selected trajectories, qualitatively corroborating Tables~\ref{tab:comparison_imu} and~\ref{tab:comparison_scr}.

\textbf{Computation time.}
The average runtimes of (1) SCR \& PnP-RANSAC, (2) fixed-lag smoothing, and (3) trajectory optimization are 62.2\,ms, 0.8\,ms, and 28.3\,ms, respectively---sufficiently low for real-time operation given the module execution periods.

\begin{figure}[!t]
	\centering         
	\begin{subfigure}[c]{0.36\linewidth}
		\centering
		\includegraphics[width=\linewidth]{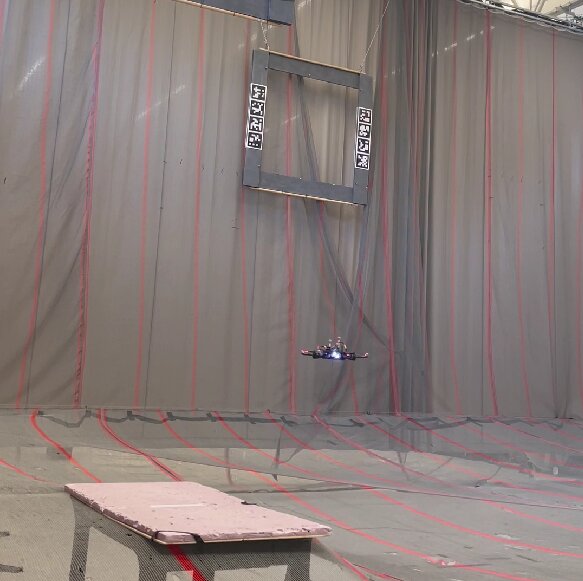}
		\caption{Physical environment}\label{fig:drone}
	\end{subfigure}
	\begin{subfigure}[c]{0.47\linewidth}
		\centering
		\includegraphics[width=\linewidth]{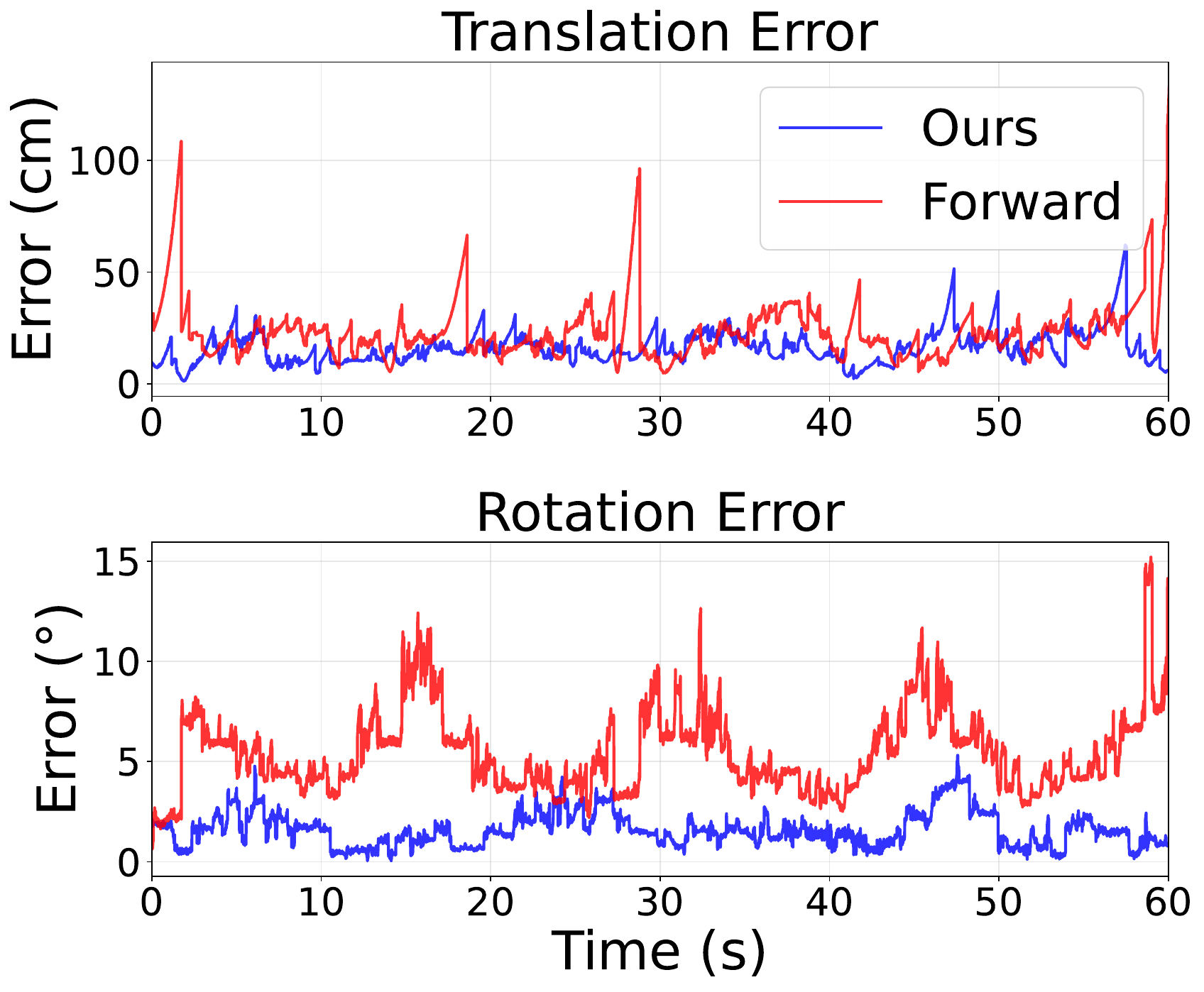}
		\caption{Translation and rotation error}\label{fig:hw_graph}
	\end{subfigure}\hfill
	\caption{Hardware-in-the-loop experiment.
	(\subref{fig:drone}) Drone and spaces for the hardware-in-the-loop experiment.
	(\subref{fig:hw_graph}) Time histories of translation error (up) and rotation error (down) for \emph{Forward} and our method.} 
	\label{fig:hiw_exp}
	\vspace{-0.25in}
\end{figure}

\subsection{Hardware-in-the-loop Experiment}
This experiment validates system-level integration under close-to-real conditions rather than re-benchmarking planners, so we compare only against \emph{Forward} as a representative reference.
The quadrotor follows a rose-petal trajectory, $r = 3\cos(2\theta)$, for three laps.

Translational RMSE and mean error of the fused pose dropped from 29.7\,cm and 24.6\,cm to 22.7\,cm and 20.0\,cm, respectively, while rotational RMSE and mean error fell from 6.0$^\circ$ and 5.6$^\circ$ to 2.3$^\circ$ and 2.0$^\circ$.
Fig.~\ref{fig:hw_graph} shows that our error remains consistently below that of \emph{Forward} over a representative lap, confirming the method's feasibility under close-to-real-world conditions.

%% file: section/conclusion.tex
This paper introduces a receding-horizon, perception-aware trajectory generation framework using evidential learning-based scene coordinate regression (SCR).
Our algorithm consistently orients the camera toward scene coordinates with low uncertainty and produces drift-free, high-accuracy poses in real-time.
Experiments demonstrate that our framework outperforms baseline methods in localization accuracy and is suitable for hardware-in-the-loop flight tests.
Our baselines share the same SCR pipeline to isolate the contribution of uncertainty-aware yaw optimization; benchmarking against VIO- and SfM-based planners is left to future work, along with obstacle avoidance and robustness to appearance changes such as lighting variations and object rearrangements.